\pdfoutput=1
\documentclass[journal]{IEEEtran}

\usepackage{amsmath,amssymb,amsfonts}
\usepackage{graphicx}
\usepackage{booktabs}
\usepackage{cite}
\usepackage{capt-of}
\usepackage{hyperref}

\title{Adaptive Traversal Tubes for Posture-Aware Navigation of Bipedal Humanoids}

\author{
  Zijian Ge ~\IEEEmembership{Member, IEEE}
  \thanks{This is a methods-oriented preprint. Quantitative experimental evaluation is ongoing and will be updated in a revised version.
  Corresponding author: Zijian Ge (e-mail: gezijian1001@gmail.com).
  }
}

\begin{document}
\maketitle

\begin{abstract}
Bipedal humanoids can change their effective footprint by adjusting body yaw, enabling both narrow-passage traversal and socially aware yielding.
This paper presents a posture-aware navigation framework centered on \emph{Adaptive Traversal Tubes} (ATT), \emph{Interval Velocity--Heading} (IVH) control, and a sidestep-enhanced AMP locomotion prior.
ATT represents a behavior-feasible corridor width field along a reference path, unifying geometric clearance and social comfort into a continuous signal.
IVH maps this width to a continuous frontal-to-sidestep posture schedule and omnidirectional velocity commands.
A velocity-conditioned AMP policy, trained with GVHMR-retargeted frontal/sidestep motions and coupled \((v_x,v_y)\) tracking rewards, is designed to improve transition stability over a generic locomotion controller while remaining interchangeable with the default onboard policy.
This preprint presents the full method; a planned evaluation protocol on a Unitree G1 is outlined, with quantitative results to follow.
Demo videos of MuJoCo simulation and real-robot deployment are available at:
\url{https://github.com/lborogzj997/Humanoid-Navigation-in-Crowds-and-Narrow-Spaces}.
\end{abstract}


\section{Introduction}
\label{sec:intro}

\begin{figure}[!t]
  \centering
  \includegraphics[width=\columnwidth]{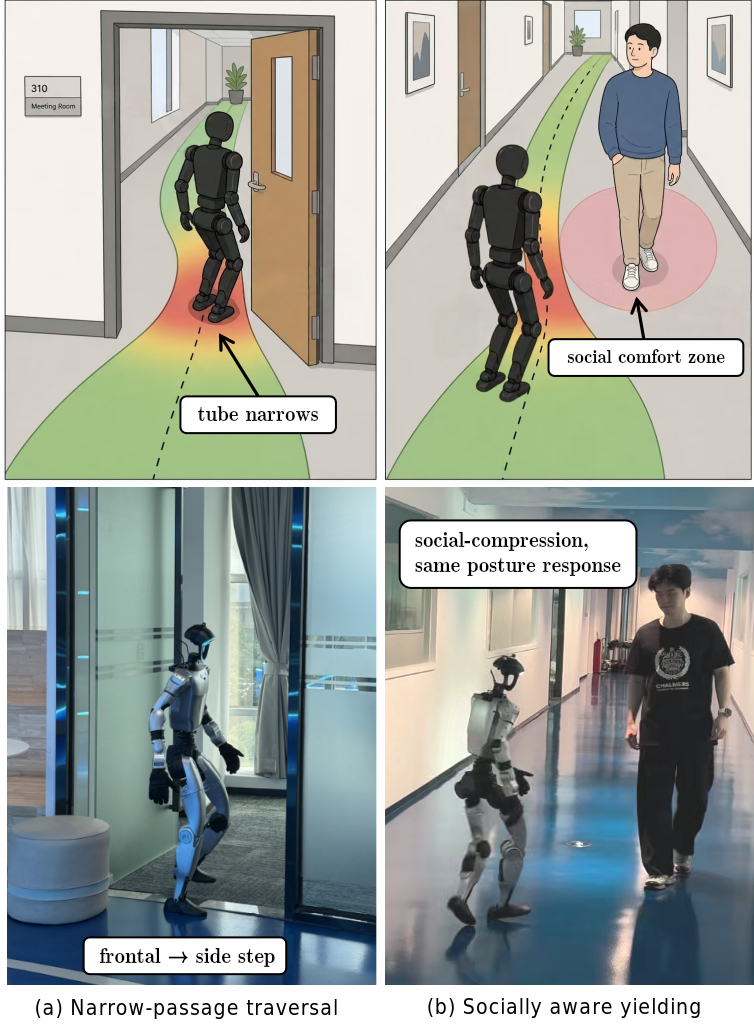}
  \caption{Overview of the proposed idea. Top: ATT width field (green$\,\to\,$red as the tube narrows) for a geometric squeeze-through (left) and social comfort compression (right). Bottom: corresponding real-robot behaviors on a Unitree G1, where IVH schedules a frontal-to-sidestep posture response.}
  \label{fig:teaser}
\end{figure}

Indoor service settings---offices, hospitals, apartments, and retail floors---are built around the human body:
doorways, corridors, and furniture clearances are often only marginally wider than a person's shoulders, while the same spaces are shared with pedestrians who expect socially acceptable distances and yielding~\cite{Kruse2013HumanAware,RiosMartinez2015Proxemics,Francis2025SocialNavPrinciples}.
Bipedal humanoids are uniquely well matched to this ecology.
Unlike wheeled platforms, whose chassis footprint is essentially fixed in the horizontal plane, a humanoid can reshape its \emph{effective} passage width by changing body yaw and locomotion style---walking face-forward in open space, then rotating into a lateral (sidestep) posture when the corridor tightens~\cite{Kanehiro2004NarrowSpaces,Kumagai2018NarrowLocomotion}.
The same posture change also supports socially polite yielding: by presenting a thinner frontal profile and shifting sideways, the robot can open a lane for a person without abandoning the intended route.
This combination of geometric squeeze-through and socially aware lateral yielding is difficult to realize with classical wheeled navigation stacks, which typically inflate a disk or polygon footprint and either declare a passage infeasible or merely slow down~\cite{Macenski2020Nav2}.

Realizing that potential, however, exposes several \emph{general challenges} that cut across applications.
First, \textbf{posture-dependent footprint ambiguity}: the clearance that matters is not raw wall-to-wall distance, but whether the robot's current (and scheduled) body orientation makes the passage behaviorally feasible.
A frontal shoulder width that is safe in a hallway may be too large for a door, while the same door may be passable in a sidestep posture; treating the robot as a fixed disk therefore both over-rejects and under-explains failures~\cite{Kanehiro2004NarrowSpaces,Kuffner2003Footstep}.
Second, \textbf{unified geometric and social constraints}: in human environments, free space is further compressed by comfort and proxemic expectations~\cite{Kruse2013HumanAware,RiosMartinez2015Proxemics}.
Prior social-navigation work---including our own congestion-aware and spatial--temporal routing methods for mobile robots~\cite{Ge2023CongestionAware,Ge2024PPUM,Ge2025SocialRouting}---has largely addressed crowd-aware \emph{path selection} and speed adaptation for platforms with essentially fixed footprints.
For bipeds, social compression should instead trigger the same continuous posture response as a physical bottleneck: earlier sidestepping and lateral yielding, not only detours or slowdowns.
Third, \textbf{navigation-layer posture scheduling under locomotion interfaces}: modern humanoids often expose omnidirectional velocity commands to a learned or commercial locomotion policy.
Contact-level footstep and whole-body planners can traverse tight spaces~\cite{Kuffner2003Footstep,Kumagai2018NarrowLocomotion}, but their computational focus is on footholds and balance rather than on an explicit, tunable mapping from corridor or social width to body yaw.
Conversely, wheeled stacks such as Nav2~\cite{Macenski2020Nav2} provide mature map-based planning yet lack a first-class notion of frontal-to-sidestep posture.
The general concern is therefore not merely find a collision-free path, but schedule posture and omnidirectional velocity so that the path remains behavior-feasible as clearance and social comfort vary along it.

This paper argues that posture should be elevated to the navigation layer through a shared width signal~(Fig.~\ref{fig:teaser}).
We introduce \emph{Adaptive Traversal Tubes} (ATT): a path-parameterized width field that represents an \emph{effective passage width} along a reference path for frontal walking decisions, rather than the raw gap between walls.
Far from obstacles this width saturates at a nominal upper bound \(w_{\mathrm{nom}}\); it shrinks when geometric clearance or social comfort tightens, thereby unifying static squeeze-through and dynamic yielding in one continuous scalar signal.
We then introduce \emph{Interval Velocity--Heading} (IVH) tracking, which maps the tube width to a continuous frontal-to-sidestep posture schedule and the corresponding omnidirectional velocity and heading commands for bipedal locomotion.
To execute these commands stably, we further train a sidestep-enhanced AMP locomotion prior from custom human videos (GVHMR capture and GMR retargeting), with task rewards that explicitly couple frontal and lateral velocity tracking during yaw transitions.
ATT and IVH remain compatible with a generic onboard controller; the enhanced prior mainly improves transition smoothness and lateral tracking when IVH requests nonzero \((v_x,v_y)\) jointly.

\paragraph{Contributions.}
\begin{enumerate}
  \item \textbf{ATT:} a behavior-feasible traversal tube along a reference path that unifies static geometric clearance and dynamic social compression into one posture-relevant width field.
  \item \textbf{IVH:} a continuous frontal--sidestep posture schedule driven by the tube width, producing omnidirectional velocity and heading commands for bipedal locomotion.
  \item \textbf{Sidestep-enhanced AMP prior:} a velocity-conditioned locomotion policy trained with frontal/sidestep reference motions and coupled \((v_x,v_y)\) tracking objectives, intended to improve stability of frontal--sidestep transitions relative to a generic locomotion controller while remaining drop-in interchangeable with it.
  \item \textbf{Evaluation protocol:} a planned simulation and real-robot comparison on a Unitree G1 against a frontal-only navigation baseline, to be reported in an updated version of this preprint.
\end{enumerate}

\section{Related Work}
\label{sec:related}

We organize prior work on bipedal humanoid navigation into three strands: contact-level footstep and whole-body control, learning-based locomotion and tightly coupled navigation policies, and geometry-/social-aware mid-level navigation.
Across these strands, a recurring gap is an explicit, lightweight interface that schedules frontal-to-sidestep posture from a shared geometric--social width signal while remaining compatible with an existing stable locomotion stack.

\paragraph{Footstep planning with WBC/MPC.}
Classical humanoid navigation plans discrete footholds (or contact sequences) and tracks them with walking-pattern generation, whole-body control (WBC), or model predictive control (MPC), while regulating CoM/ZMP or related balance quantities~\cite{Kuffner2003Footstep,Scianca2020ISMPC}.
This line of work is particularly strong when contact timing, rough terrain, and balance certificates dominate, and it has long recognized that humanoids can alter locomotion style to pass narrow spaces~\cite{Kanehiro2004NarrowSpaces,Kumagai2018NarrowLocomotion}.
Subsequent work further develops real-time pattern generation with nonlinear MPC and obstacle avoidance~\cite{Naveau2017RAL}, as well as efficient contact-surface selection via sparse optimization for online footstep planning~\cite{Song2021SL1M}.
Recent formulations continue to push whole-body feasibility in footstep MPC~\cite{Scianca2020ISMPC,Belvedere2024JISMPC}.
Their computational focus, however, remains on \emph{where to step} and \emph{how to stay balanced}.
Body yaw as a continuous navigational degree of freedom---used to reshape the effective frontal footprint for squeeze-through or yielding---is typically not exposed as a tunable navigation-layer schedule; when sidestepping appears, it is often a byproduct of contact optimization rather than an explicit posture policy.
For indoor corridor and social settings where a reliable velocity-conditioned locomotion controller already exists, solving contact-level WBC/MPC online can also be heavier than necessary for the posture-scheduling decision we target.

\paragraph{Learning-based locomotion and tightly coupled navigation.}
Reinforcement learning (RL) has substantially advanced bipedal locomotion, from motion-prior stylization~\cite{Peng2021AMP} and adaptive CPG-style bipedal controllers~\cite{Li2024AICPG} to real-world humanoid walking policies~\cite{Radosavovic2024SciRob}.
Building on such controllers, recent work couples navigation more tightly with learning: emotion-aware social navigation for bipeds~\cite{Zhu2026EmoBipedNav}, end-to-end LiDAR-to-motor safe/comfortable policies~\cite{Wang2025SafeHumanoidPolicy}, physics-guided narrow-path traversal~\cite{Huang2025NarrowPathRL}, and navigation priors learned from human walking data~\cite{Wang2026EgoNav}.
These methods can discover agile behaviors with limited hand-designed structure, but they typically entail costly training, careful reward or constraint design, and nontrivial sim-to-real transfer when perception, posture transitions, and balance are learned jointly.
Retargeting to a new map stack, planner, or locomotion backend often requires substantial retraining rather than local parameter retuning.
We keep ATT+IVH modular with respect to the locomotion backend, and additionally train a sidestep-enhanced AMP prior~(Section~\ref{sec:loco}) that better matches IVH's coupled \((v_x,v_y)\) and yaw-transition commands than a generic walking controller.

\paragraph{Geometry-aware and social bipedal navigation.}
A complementary trend keeps learned (or model-based) locomotion in the inner loop and adds mid-level planners or safety filters for cluttered scenes: hierarchical task-and-motion planning for safe bipedal navigation under partial observability~\cite{Shamsah2023TRO}, Koopman-lifted MPC for narrow passages~\cite{Kim2024KoopmanNav}, geometry-aware CBF--MPC safety filters that can reorient a humanoid through tight gaps~\cite{Bena2025GeomAwareSafety}, and related CBF safety layers on learned humanoid controllers~\cite{Yang2025SHIELD}.
For human-shared spaces, social bipedal navigation has begun to incorporate proxemics and pedestrian prediction into footstep MPC~\cite{Shamsah2025SZNMPC}, while wheeled social planners increasingly inject context or language-model cues into navigation costs~\cite{Stefanini2024ContextMPC,Song2025VLMSocialNav}.
Classical wheeled stacks and social-navigation surveys, however, largely assume a fixed footprint and solve crowding by detours, inflation, or slowing down~\cite{Macenski2020Nav2,Kruse2013HumanAware,RiosMartinez2015Proxemics,Francis2025SocialNavPrinciples}.
These approaches are closest in spirit to our setting, yet most still either optimize contact sequences under social costs, or filter velocities for collision safety, without an explicit behavior-feasible width field that unifies geometric squeeze-through and social comfort into one continuous posture schedule.

\paragraph{Positioning of this work.}
Relative to footstep--WBC/MPC pipelines, ATT+IVH do not aim to replace contact-level optimality; they provide a lighter navigation interface that schedules frontal--sidestep posture from tube width and leaves balance to the locomotion layer.
Relative to end-to-end or tightly coupled RL navigation, we trade joint training flexibility for modularity: the same ATT/IVH block can sit atop different velocity-conditioned locomotion controllers with limited retuning, and the width field remains visualizable and parameter-interpretable.
Relative to geometry-aware safety filters and social footstep MPC, we emphasize a shared path-parameterized width signal for both geometric and social compression, mapped by IVH into omnidirectional velocity--heading commands.
In short, whenever a bipedal platform already offers stable omnidirectional locomotion, ATT+IVH add posture-aware passage and yielding without requiring a full contact planner or a navigation policy retrained from scratch.

\begin{figure*}[!t]
  \centering
  \includegraphics[width=\textwidth]{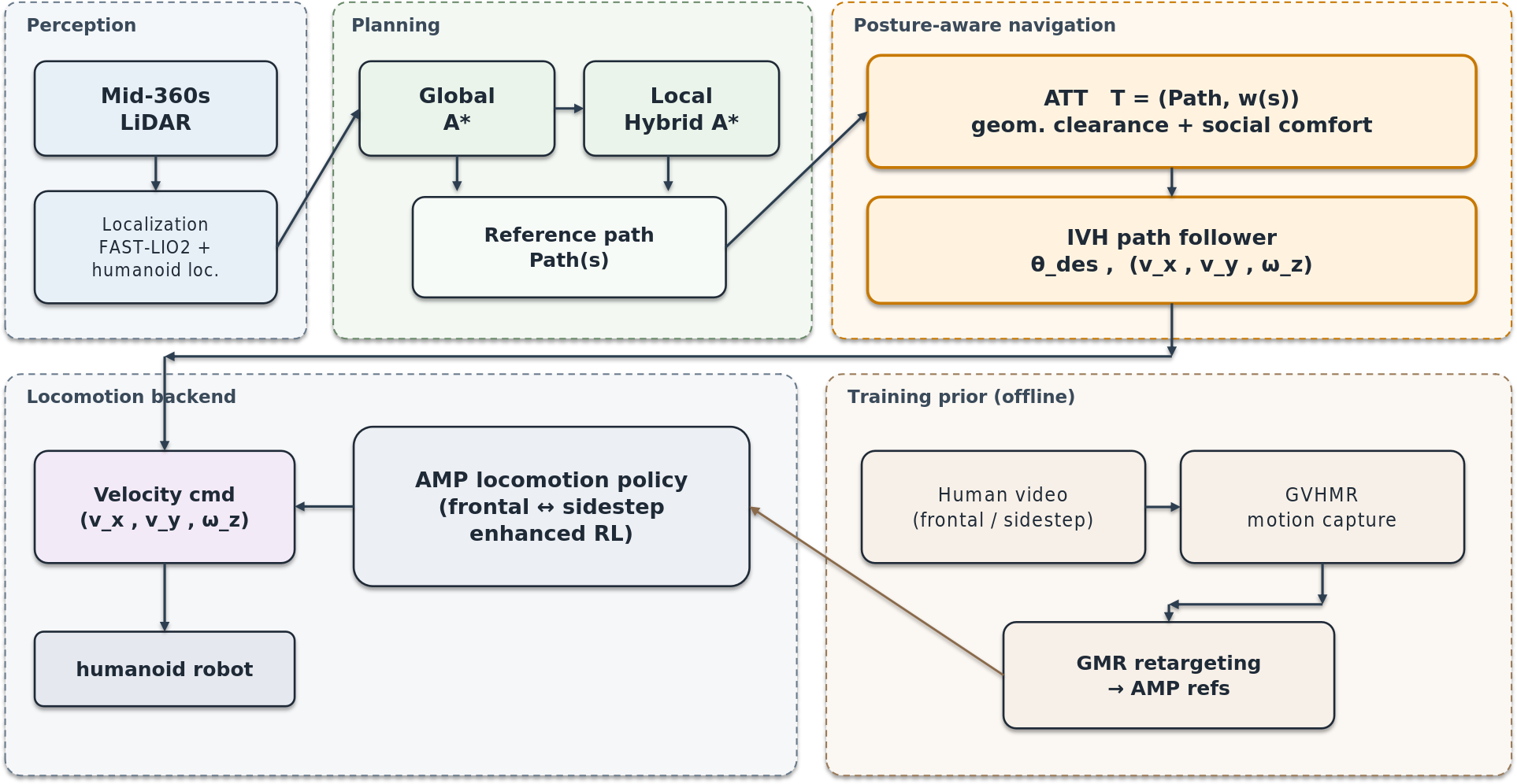}
  \caption{System overview. Mid-360 LiDAR localization supports global A* and local Hybrid~A* planning.
  ATT builds a behavior-feasible width field \(w(s)\) along the reference path from geometric clearance and social comfort;
  IVH maps \(w(s)\) to a frontal--sidestep heading schedule and body-frame \((v_x,v_y,\omega_z)\) for a sidestep-enhanced AMP locomotion policy.
  Offline, GVHMR+GMR provide frontal/sidestep reference motions for AMP training; a generic onboard controller remains a compatible fallback.}
  \label{fig:system}
\end{figure*}

\section{Method}
\label{sec:method}

\subsection{System Overview}
\label{sec:overview}

Fig.~\ref{fig:system} summarizes the stack.
Onboard Mid-360 LiDAR with FAST-LIO2-style localization maintains the robot pose on a prior map.
A global A* planner produces a long-horizon reference path; a local Hybrid~A* module can refine the path in the near field under updated costmaps.
ATT attaches a scalar width field \(w(s)\) to this path: the effective frontal passage width under geometry and, when available, social proximity.
IVH then tracks the path while scheduling a continuous frontal-to-sidestep heading and emitting body-frame omnidirectional commands \((v_x,v_y,\omega_z)\).
These commands are executed by a velocity-conditioned locomotion policy: by default we deploy a sidestep-enhanced AMP prior (Section~\ref{sec:loco}); a generic onboard controller can be substituted with the same IVH interface, typically at the cost of less stable frontal--sidestep transitions.

\subsection{Adaptive Traversal Tube (ATT)}
\label{sec:att}

\paragraph{Tube definition.}
Let the planner output a reference path as a polyline with points \(\mathbf{p}(s)\in\mathbb{R}^{2}\), arc length \(s\in[0,L]\), unit tangent \(\mathbf{t}(s)\), and left unit normal \(\mathbf{n}(s)\).
ATT augments this path with a scalar field \(w(s)>0\) and defines the tube
\begin{equation}
\label{eq:att_tube}
\mathcal{T}=\bigl(\{\mathbf{p}(s)\}_{s\in[0,L]},\; w(s)\bigr).
\end{equation}
Geometrically, the left/right tube edges are \(\mathbf{p}(s)\pm\tfrac{1}{2}w(s)\,\mathbf{n}(s)\).
Physically, \(w(s)\) is the \emph{effective passage width} used later by IVH for posture scheduling:
it answers ``how wide is the frontal walkway the robot should assume at \(s$?'', not ``what is the Euclidean gap between the two nearest walls''.
In free space far from obstacles and agents, \(w(s)\) saturates at a fixed upper bound \(w_{\mathrm{nom}}\) (defined below);
near walls or people, \(w(s)\) decreases so that IVH can request a stronger sidestep.
We first construct a geometry-only width \(w_{\mathrm{static}}(s)\), then optionally take the pointwise minimum with a social term to obtain the final \(w(s)\).

\paragraph{Static geometric width.}
From an occupancy grid (global map and/or local costmap), compute the Euclidean distance transform and define the clearance
\begin{equation}
\label{eq:clearance}
c(s)
\;=\;
\mathrm{dist}\bigl(\mathbf{p}(s),\,\mathcal{O}\bigr)
\;\ge\;0,
\end{equation}
i.e., the distance from the path sample \(\mathbf{p}(s)\) to the nearest occupied cell set \(\mathcal{O}\).
Interpreting \(c(s)\) as a half-width estimate of free space about the centerline, a raw bilateral width would be \(2c(s)\).
We convert clearance into a bounded tube width by
\begin{equation}
\label{eq:w_static}
w_{\mathrm{static}}(s)
=
\mathrm{clip}\!\bigl(2\,c(s)-m_{\mathrm{safe}},\;
w_{\min},\;
w_{\mathrm{nom}}\bigr),
\end{equation}
where \(\mathrm{clip}(x;a,b)=\min\{\max\{x,a\},b\}\) and the three scalars have direct physical roles:
\begin{itemize}
  \item \(m_{\mathrm{safe}}>0\): a fixed length subtracted from \(2c(s)\) to leave margin for map noise, localization error, and the robot's body thickness. Larger \(m_{\mathrm{safe}}\) yields a narrower tube (more conservative).
  \item \(w_{\mathrm{nom}}>0\): the \emph{maximum} tube width allowed by ATT. When the robot is far from obstacles (\(2c(s)-m_{\mathrm{safe}}\ge w_{\mathrm{nom}}\)), we set \(w_{\mathrm{static}}(s)=w_{\mathrm{nom}}\) instead of letting the width grow without bound. This bound is intentional: IVH only needs to know that space is ``wide enough for frontal walking'', not the true open-area size.
  \item \(w_{\min}\in(0,w_{\mathrm{nom}}]\): the \emph{minimum} tube width. It keeps \(w_{\mathrm{static}}\) strictly positive in extreme bottlenecks so that downstream ratios and controllers remain well-defined.
\end{itemize}
After \eqref{eq:w_static}, we apply a short sliding average along \(s\) and, near local minima of the width, an optional blend of length \(\ell_{\mathrm{blend}}\) so that \(w_{\mathrm{static}}(s)\) does not jump abruptly at doorways or corners.
If both a global map and a rolling local costmap are available, clearances inside a forward horizon of the robot are taken from the local map, and beyond that horizon from the global map, with a smooth blend at the junction.

\paragraph{Social compression.}
When nearby agents are detected, ATT reduces the tube width while keeping the same centerline \(\mathbf{p}(s)\).
Let \(d(s)\) be the distance from \(\mathbf{p}(s)\) to the nearest agent, and let \(r_{\mathrm{a}}>0\) be that agent's radius (or a comfort radius).
The residual free distance to the agent's comfort boundary is \(\max(0,\,d(s)-r_{\mathrm{a}}-m_{\mathrm{safe}})\); doubling it yields a social width analogous to \(2c(s)-m_{\mathrm{safe}}\).
The effective width used by IVH is then
\begin{equation}
\label{eq:w_eff}
\begin{aligned}
w(s)
=\min\Bigl(
&w_{\mathrm{static}}(s),\\
&\mathrm{clip}\bigl(
2\max(0,\,d(s)-r_{\mathrm{a}}-m_{\mathrm{safe}}),\,
w_{\min},\,w_{\mathrm{nom}}
\bigr)
\Bigr).
\end{aligned}
\end{equation}
Hence a geometric squeeze and a social approach act through the \emph{same} scalar \(w(s)\): both decrease \(w(s)\) and later raise the same IVH sidestep ratio.
If no agent is present, the social term is inactive and \(w(s)=w_{\mathrm{static}}(s)\); in particular, far from obstacles one has \(w(s)=w_{\mathrm{nom}}\).

\subsection{Interval Velocity--Heading (IVH)}
\label{sec:ivh}

\paragraph{Control objective.}
IVH sits between ATT and the locomotion backend.
Its input is the tube width \(w(s)\) from Section~\ref{sec:att}; its output is a body-frame command \((v_x,v_y,\omega_z)\) that a velocity-conditioned walking policy can track.
The design follows a fixed chain:
\[
w(s)
\;\rightarrow\;
\rho(s)
\;\rightarrow\;
\theta_{\mathrm{des}}(s)
\;\rightarrow\;
(v_x,v_y,\omega_z).
\]
Here \(\rho(s)\in[0,1]\) is a \emph{sidestep ratio} (how much the robot should face sideways), and \(\theta_{\mathrm{des}}(s)\) is the desired body yaw.
The name ``interval'' refers to the fact that posture is not a binary frontal/sidestep switch: as \(w(s)\) varies between two body-size thresholds, \(\rho\) moves continuously inside \([0,1]\).

\paragraph{Sidestep ratio.}
Let \(w_{\mathrm{sh}}\) be the robot's nominal frontal width (roughly shoulder-to-shoulder), and \(w_{\mathrm{dp}}\) its body depth (front-to-back), with \(w_{\mathrm{dp}}<w_{\mathrm{sh}}\).
These two lengths are the natural thresholds for posture:
\begin{itemize}
  \item if the tube is at least as wide as the shoulders (\(w(s)\ge w_{\mathrm{sh}}\)), the robot can walk face-forward;
  \item if the tube is no wider than the body depth (\(w(s)\le w_{\mathrm{dp}}\)), a full sidestep is required to fit;
  \item between these values, the robot should rotate gradually from frontal to sidestep.
\end{itemize}
We encode this with
\begin{equation}
\label{eq:rho}
\rho(s)
=
\begin{cases}
0, & w(s)\ge w_{\mathrm{sh}},\\[2pt]
1, & w(s)\le w_{\mathrm{dp}},\\[2pt]
1-S\!\bigl(w(s);\,w_{\mathrm{dp}},\,w_{\mathrm{sh}}\bigr), & \text{otherwise},
\end{cases}
\end{equation}
so that \(\rho=0\) means frontal walking, \(\rho=1\) means full sidestep, and intermediate \(\rho\) means a partial yaw offset.
The map \(S(x;a,b)\) first normalizes \(x\) onto the unit interval,
\[
t=\mathrm{clip}\!\bigl((x-a)/(b-a),\,0,\,1\bigr),
\]
and then applies the standard cubic smoothstep
\begin{equation}
\label{eq:smoothstep}
S(t)=t^{2}(3-2t).
\end{equation}
This cubic is the unique polynomial of degree \(\le 3\) with \(S(0)=0\), \(S(1)=1\), and \(S'(0)=S'(1)=0\).
The zero endpoint slopes matter in practice: \(\rho(s)\) (and therefore \(\theta_{\mathrm{des}}\)) stays \(C^{1}\) when \(w(s)\) crosses \(w_{\mathrm{dp}}\) or \(w_{\mathrm{sh}}\), so the robot does not get a kinked yaw schedule at the start or end of a narrow segment.

\paragraph{Desired heading.}
Let \(\psi(s)=\mathrm{atan2}(t_y(s),t_x(s))\) be the path tangent heading, and let \(\sigma(s)\in\{-1,+1\}\) choose the sidestep side (left or right).
The desired body yaw is a path-aligned heading plus a sidestep offset proportional to \(\rho\):
\begin{equation}
\label{eq:theta}
\theta_{\mathrm{des}}(s)
=
\psi(s)+\sigma(s)\cdot\frac{\pi}{2}\cdot\rho(s).
\end{equation}
When \(\rho=0\), \(\theta_{\mathrm{des}}=\psi\) (face along the path).
When \(\rho=1\), \(\theta_{\mathrm{des}}=\psi\pm\pi/2\) (face perpendicular to the travel direction).
Side \(\sigma\) is chosen once at the onset of a narrowing segment by a minimum-yaw rule: among \(\{+\pi/2,-\pi/2\}\), pick the rotation closer to the robot's current yaw.
A hysteresis latch then holds \(\sigma\): after \(\rho\) rises above an enter threshold, \(\sigma\) is frozen until \(\rho\) falls below a lower release threshold.
This avoids left/right chatter when \(w(s)\) oscillates near \(w_{\mathrm{sh}}\) or \(w_{\mathrm{dp}}\).
Optional biases (path curvature, or the lateral side of a nearby pedestrian) can prefer one \(\sigma\) at latch time; once latched, a soft local replan does not flip the committed passing side.

\paragraph{Body-frame velocity commands.}
At each control tick we project the robot pose onto the reference path to obtain the current arc length \(s\), the signed lateral error \(e_{\mathrm{lat}}\), and lookahead samples of \(\rho\) and \(\theta_{\mathrm{des}}\).
A progress speed \(v\ge 0\) starts from a cruise value, is reduced near the goal, and is further scaled down when \(\rho\) is large (narrow tube \(\Rightarrow\) move more carefully).
The preferred planar velocity in the world frame combines forward progress along the tangent with a lateral correction toward the path:
\begin{equation}
\label{eq:vw}
\mathbf{v}_{w}
=
v\,\mathbf{t}(s)
-
k_{\mathrm{lat}}\,e_{\mathrm{lat}}\,\mathbf{n}(s).
\end{equation}
Here \(k_{\mathrm{lat}}>0\) is a lateral gain; it may be increased during mid-transition (intermediate \(\rho\) with large heading error) so that the robot stays centered while yawing.
Commands are then rotated into the body frame and closed-loop on yaw:
\begin{equation}
\label{eq:cmd}
\begin{aligned}
(v_x,v_y)
&=\mathbf{R}(\mathrm{yaw})^{\!\top}\mathbf{v}_{w},\\
\omega_z
&=\mathrm{clip}\bigl(
k_{\theta}\,\mathrm{wrap}(\theta_{\mathrm{des}}-\mathrm{yaw}),\,
\pm\omega_{\max}
\bigr),
\end{aligned}
\end{equation}
where \(k_{\theta}>0\) is a heading gain, \(\mathrm{wrap}(\cdot)\) wraps the angle error to \((-\pi,\pi]\), and \(\omega_{\max}\) limits the yaw rate.
We additionally apply rate limits and low-pass filtering on \((v_x,v_y,\omega_z)\), and small dead-zones that keep a minimum \(|v_y|\) (resp.\ \(v_x\)) once sidestep (resp.\ frontal) motion is committed, so the command does not chatter about zero during a transition.
Because both geometry and social proximity already entered through \(w(s)\rightarrow\rho(s)\), the same IVH law produces squeeze-through at doors and earlier yielding when comfort narrows the tube.

\subsection{Sidestep-Enhanced AMP Locomotion}
\label{sec:loco}

\paragraph{Motivation for a sidestep-enhanced prior.}
ATT and IVH do not balance the robot; they only emit body-frame commands \((v_x,v_y,\omega_z)\).
Any locomotion backend that tracks these commands is therefore valid, including a generic onboard controller (e.g., the Unitree G1 default policy).
In practice, generic priors are biased toward frontal walking.
They degrade when IVH asks for sustained lateral speed \(v_y\), or when \(v_x\) and \(v_y\) stay nonzero together while yaw rotates through a frontal--sidestep transition.
We therefore train a preferred backend: a \emph{sidestep-enhanced} AMP policy~\cite{Peng2021AMP}.
The change is deliberately narrow---we do not redesign navigation:
\begin{enumerate}
  \item enrich the motion library with the frontal, sidestep, and transition gaits that IVH actually requests;
  \item add a steering task that treats \emph{travel direction} and \emph{facing direction} as independent commands, so crab-walk and continuous yaw change are first-class behaviors.
\end{enumerate}

\paragraph{Reference motions.}
We record monocular videos of a human demonstrator performing the three command patterns that IVH emits:
\begin{enumerate}
  \item \textbf{Frontal:} face along the travel direction, roughly \(0.3\)--\(1.0\,\mathrm{m/s}\), on straight or gently curved paths;
  \item \textbf{Sidestep:} crab-walk with the torso about \(\pm 90^\circ\) from the travel direction, at similar speeds, for both left and right;
  \item \textbf{Transitions:} continuous yaw change while translating (frontal\(\leftrightarrow\)sidestep), including slow in-place pivots and coupled \((v_x,v_y)\) progress during the turn.
\end{enumerate}
Clips last several seconds.
We keep roughly equal duration of frontal, left-sidestep, right-sidestep, and transition footage so that the AMP discriminator is not dominated by frontal gait.
Each video is lifted to world-grounded human motion with GVHMR~\cite{Shen2024GVHMR} (SMPL-X), then retargeted to the Unitree G1 with GMR~\cite{Araujo2025GMR} (\(\texttt{unitree\_g1}\) configuration).
Retargeted clips are converted to the AMP motion format and replayed on the simulated G1 to check foot contact, root height, and self-penetration.
Clips with large retargeting artifacts are discarded or re-recorded; we do not rely on RL to ``fix'' bad demonstrations.

\paragraph{Training objective.}
Training follows AMP~\cite{Peng2021AMP}: a policy \(\pi\) is optimized with PPO, while a discriminator \(D\) supplies a style reward \(r_{\mathrm{disc}}\) that pushes the robot's state transitions toward the retargeted library.
To make \(\pi\) commandable like IVH, each episode also specifies an independent steering command
\[
\bigl(\mathbf{d}_t,\,v^\star,\,\mathbf{f}\bigr),
\]
where \(\mathbf{d}_t\in\mathbb{R}^2\) is a unit travel direction in the world frame, \(v^\star\ge 0\) is the desired speed, and \(\mathbf{f}\in\mathbb{R}^2\) is a unit facing direction.
This matches IVH's posture schedule directly:
\begin{itemize}
  \item frontal walking: \(\mathbf{f}\approx\mathbf{d}_t\) (face along travel);
  \item pure sidestep: \(\mathbf{f}\perp\mathbf{d}_t\) (face sideways while moving laterally);
  \item intermediate postures: the angle between \(\mathbf{f}\) and \(\mathbf{d}_t\) lies in \((0,\pi/2)\).
\end{itemize}
Let \(\mathbf{v}\) be the measured planar root velocity and \(\mathbf{h}\) the robot's planar heading vector.
The task reward is
\begin{equation}
\label{eq:loco_task}
\begin{aligned}
r_{\mathrm{task}}
&=
w_v\,\exp\!\bigl(-\alpha_v\,\lVert v^\star\mathbf{d}_t-\mathbf{v}\rVert_2^2\bigr)
+
w_f\,\max\bigl(0,\,\mathbf{f}^\top\mathbf{h}\bigr),
\end{aligned}
\end{equation}
with a clear split of roles:
\begin{itemize}
  \item velocity term: reward matching the commanded planar velocity \(v^\star\mathbf{d}_t\); \(\alpha_v>0\) sets how sharply the reward falls when tracking error grows (larger \(\alpha_v\) \(\Rightarrow\) stricter speed tracking);
  \item facing term: reward aligning heading \(\mathbf{h}\) with \(\mathbf{f}\) via a clipped cosine \(\mathbf{f}^\top\mathbf{h}\) (no credit for facing the opposite way);
  \item \(w_v,w_f>0\): relative weights of speed tracking versus facing (defaults \(w_v=0.7\), \(w_f=0.3\)).
\end{itemize}
If the projected speed along \(\mathbf{d}_t\) is negative, we zero the velocity term so that the policy cannot score well by walking backward during sidestep.
The total reward mixes task and style,
\begin{equation}
\label{eq:loco_total}
r = \lambda_{\mathrm{task}}\,r_{\mathrm{task}} + \lambda_{\mathrm{disc}}\,r_{\mathrm{disc}},
\end{equation}
with default \(\lambda_{\mathrm{task}}=\lambda_{\mathrm{disc}}=0.5\) and \(\alpha_v=0.5\).
If indoor G1 speeds are slower than the human clips, \(\alpha_v\) is lowered so that the exponential velocity reward does not become overly peaked.

\paragraph{Curriculum and deployment.}
Steering commands are resampled every \(4\)--\(7\,\mathrm{s}\) inside each episode.
Let \(\phi\) be the signed angle from \(\mathbf{d}_t\) to \(\mathbf{f}\).
Training uses a short curriculum over \(\phi\):
\begin{itemize}
  \item Stage~A (optional): pure style learning with \(\lambda_{\mathrm{task}}=0\) until \(r_{\mathrm{disc}}\) is stable;
  \item Stage~B: discrete postures \(\phi\in\{0,\pm\pi/2\}\) with equal probability (frontal and left/right sidestep), \(v^\star\sim\mathcal{U}[0.2,0.8]\,\mathrm{m/s}\);
  \item Stage~C: continuous \(\phi\sim\mathcal{U}[-\pi/2,\pi/2]\), plus occasional segments where \(\mathbf{d}_t\) and \(\mathbf{f}\) rotate together so that yaw rates resemble IVH transitions.
\end{itemize}
Observations append a body-frame encoding of \((\mathbf{d}_t,v^\star,\mathbf{f})\) to the proprioceptive AMP state.
At deployment, IVH's \((v_x,v_y,\omega_z)\) is converted into an equivalent \((\mathbf{d}_t,v^\star,\mathbf{f})\) (or the policy is trained directly on body-frame velocity commands---an equivalent conditioning).
We train in large-batch parallel simulation (on the order of \(4096\) environments) on the G1 asset, with \(10\,\mathrm{s}\) episodes, fall terminations, and standard AMP discriminator settings (replay buffer, gradient penalty).
After convergence, the actor is exported behind the same SDK bridge as the default G1 controller, so ATT and IVH stay unchanged.

\paragraph{Backend swap.}
Because IVH already outputs \((v_x,v_y,\omega_z)\), replacing the enhanced AMP prior with a generic onboard policy is a backend swap only.
ATT+IVH do not require the enhanced prior for correctness; the generic controller remains a valid fallback for validation, typically with larger heading/velocity tracking error and less stable balance during frontal--sidestep transitions.
The enhanced prior is preferred when transition smoothness and sustained lateral tracking matter.

\begin{figure*}[!t]
  \centering
  \includegraphics[width=\textwidth]{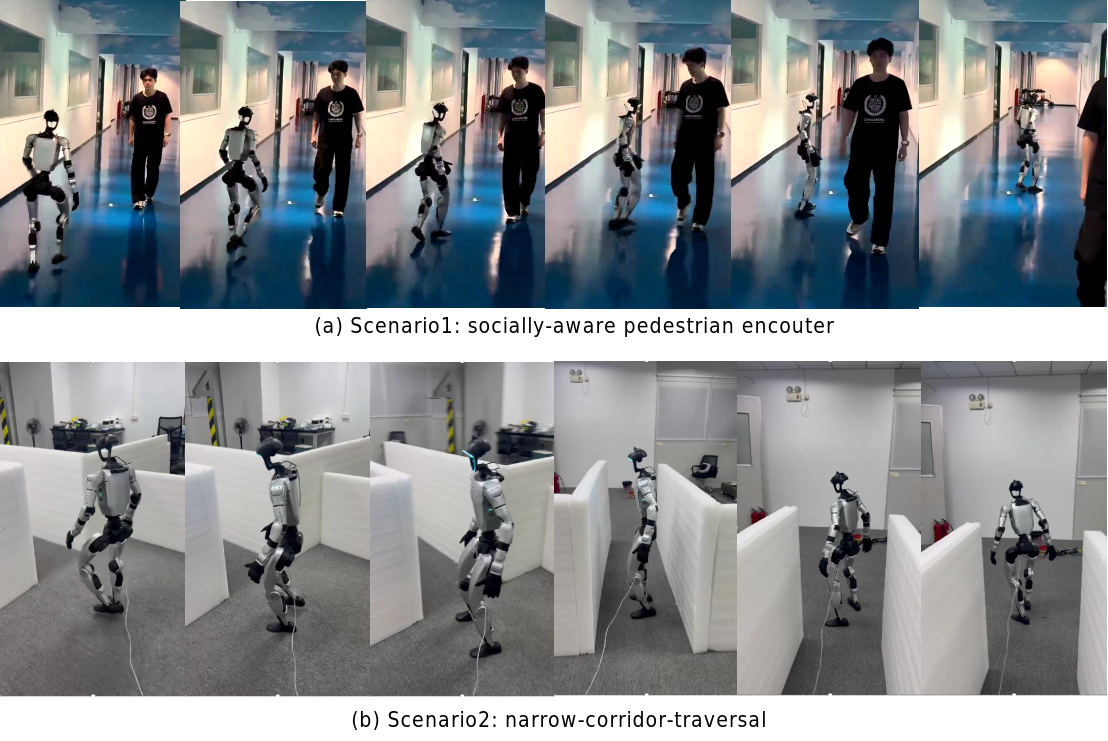}
  \caption{Qualitative illustration of the intended behaviors on a Unitree G1 (not a quantitative benchmark).
  (a)~Socially aware pedestrian encounter: comfort-space compression narrows the tube and IVH schedules a frontal-to-sidestep yield.
  (b)~Narrow-corridor traversal: geometric clearance shrinks \(w(s)\) at the gap, triggering a sidestep squeeze-through and recovery to frontal gait.
  Controlled comparisons against a frontal-only baseline are left to the evaluation protocol in Section~\ref{sec:experiments}.}
  \label{fig:scenarios}
\end{figure*}

\section{Planned Evaluation}
\label{sec:experiments}

This section specifies the evaluation protocol for an upcoming revision.
No quantitative success rates or timing statistics are claimed in the present preprint.

\subsection{Platform and Scenarios}
\label{sec:setup}

Target platform: Unitree G1 with Mid-360 LiDAR localization, map-based A*/Hybrid~A* planning, ATT+IVH, and either the sidestep-enhanced AMP prior or the default onboard locomotion backend.
Primary scenarios (Fig.~\ref{fig:scenarios}):
\begin{enumerate}
  \item \textbf{Narrow corridor / doorway:} gap width between the robot's frontal shoulder width and body depth, so that frontal-only passage is expected to fail or contact obstacles while sidestep passage remains feasible.
  \item \textbf{Social yield (optional):} an oncoming pedestrian along a still-feasible path, testing earlier tube compression and sidestep yielding without a large detour.
\end{enumerate}

\subsection{Baselines and Metrics}
\label{sec:baselines}

We plan a controlled comparison that isolates posture scheduling:
\begin{itemize}
  \item \textbf{Frontal-only (baseline):} the same localization, reference path, and locomotion backend, but with body yaw locked to the path tangent and \(v_y\!\approx\!0\) (no ATT/IVH sidestep schedule).
  \item \textbf{ATT+IVH (ours):} full width-driven frontal--sidestep scheduling as in Sections~\ref{sec:att}--\ref{sec:ivh}.
\end{itemize}
Optional ablations for a later revision include: binary \(\pm 90^\circ\) yaw without smooth \(\rho(s)\); geometry-only \(w(s)\) without social compression; and default locomotion versus the enhanced AMP prior under identical IVH commands.

Reported metrics will include success rate, collision/stuck counts, minimum obstacle clearance (and minimum interpersonal distance when applicable), and traversal time.
Transition-quality metrics (heading error, coupled \((v_x,v_y)\) tracking error) will be used when comparing locomotion backends.

\subsection{Status}
\label{sec:results}

Quantitative tables and statistical comparisons are intentionally omitted here and will be added once trials are completed.
Until then, Fig.~\ref{fig:scenarios} should be read only as a qualitative system illustration.
Accompanying simulation and Unitree~G1 demo videos are hosted at
\url{https://github.com/lborogzj997/Humanoid-Navigation-in-Crowds-and-Narrow-Spaces}.

\section{Conclusion}
\label{sec:conclusion}

We presented ATT, IVH, and a sidestep-enhanced AMP locomotion prior for posture-aware bipedal navigation: a behavior-feasible corridor width, a continuous frontal--sidestep schedule, and a velocity-conditioned gait intended to track coupled omnidirectional commands during posture transitions.
A generic onboard controller remains a compatible fallback.
This preprint focuses on the method; quantitative evaluation against a frontal-only baseline on a Unitree G1, denser social tests, and locomotion-backend ablations are planned for an updated version.

\bibliographystyle{IEEEtran}
\bibliography{refs}

\end{document}